\pdfoutput=1

\documentclass[11pt]{article}

\usepackage[]{acl2024}
\usepackage{colortbl}
\usepackage{amsfonts}
\usepackage{amsmath}
\usepackage{booktabs}
\usepackage{adjustbox}
\usepackage{graphicx}
\usepackage{arydshln}
\usepackage{float}
\usepackage{multirow}

\usepackage{times}
\usepackage{latexsym}
\usepackage{color,soul}
\usepackage[T1]{fontenc}

\usepackage[utf8]{inputenc}

\usepackage{inconsolata}
\usepackage{microtype}

\usepackage{silence}
\usepackage{xcolor}

\usepackage{enumitem}
\usepackage{array,booktabs,ragged2e}
\usepackage{makecell}
\usepackage{relsize}
\usepackage{tablefootnote}
\usepackage{caption, subcaption}
\usepackage[flushleft]{threeparttable}
\usepackage[ruled]{algorithm2e}
\usepackage{listings}
\usepackage{upquote}

\usepackage{tipa}

\usepackage{pifont} 
\usepackage{scalerel,xparse}
\usepackage{mathtools}
\usepackage{amsmath,amssymb,amsfonts}

\usepackage{todonotes}
\usepackage{tcolorbox}

\definecolor{mygreen}{HTML}{38761D}
\definecolor{myred}{HTML}{990000}
\definecolor{myblue}{HTML}{2F5596}
\definecolor{mylightblue}{HTML}{0073B7}
\definecolor{myprefixgreen}{HTML}{009E73}
\definecolor{myorange}{HTML}{D55E00}
\definecolor{mygray}{HTML}{999999}
\definecolor{myyellow}{HTML}{DAA520}
\definecolor{mydarkgoldenrod}{HTML}{B8860B}
\definecolor{mypurple}{HTML}{6F42C1}
\definecolor{myolivegreen}{HTML}{556B2F}
\definecolor{mydarkslateblue}{HTML}{483D8B}
\definecolor{lightcoral}{HTML}{F08080}
\definecolor{lightslategray}{HTML}{778899}

\newcommand{\improve}[1]{{\color[HTML]{CB4335} $\uparrow$}}
\newcommand{\worsen}[1]{{\color[HTML]{2E86C1} $\downarrow$}}
\newcommand{\best}[1]{{\color[HTML]{CB4335} $\checkmark$}}

\newcommand{\treelogo}{\raisebox{5pt}{\includegraphics[scale=0.050]{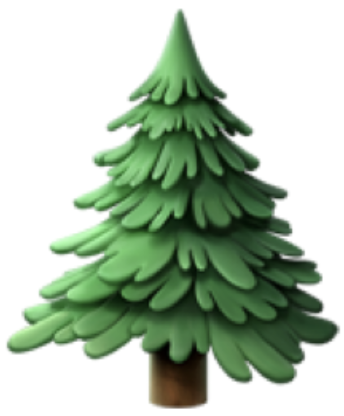}}}
\newcommand{\hlogo}{\raisebox{3.4pt}{\includegraphics[scale=0.0085]{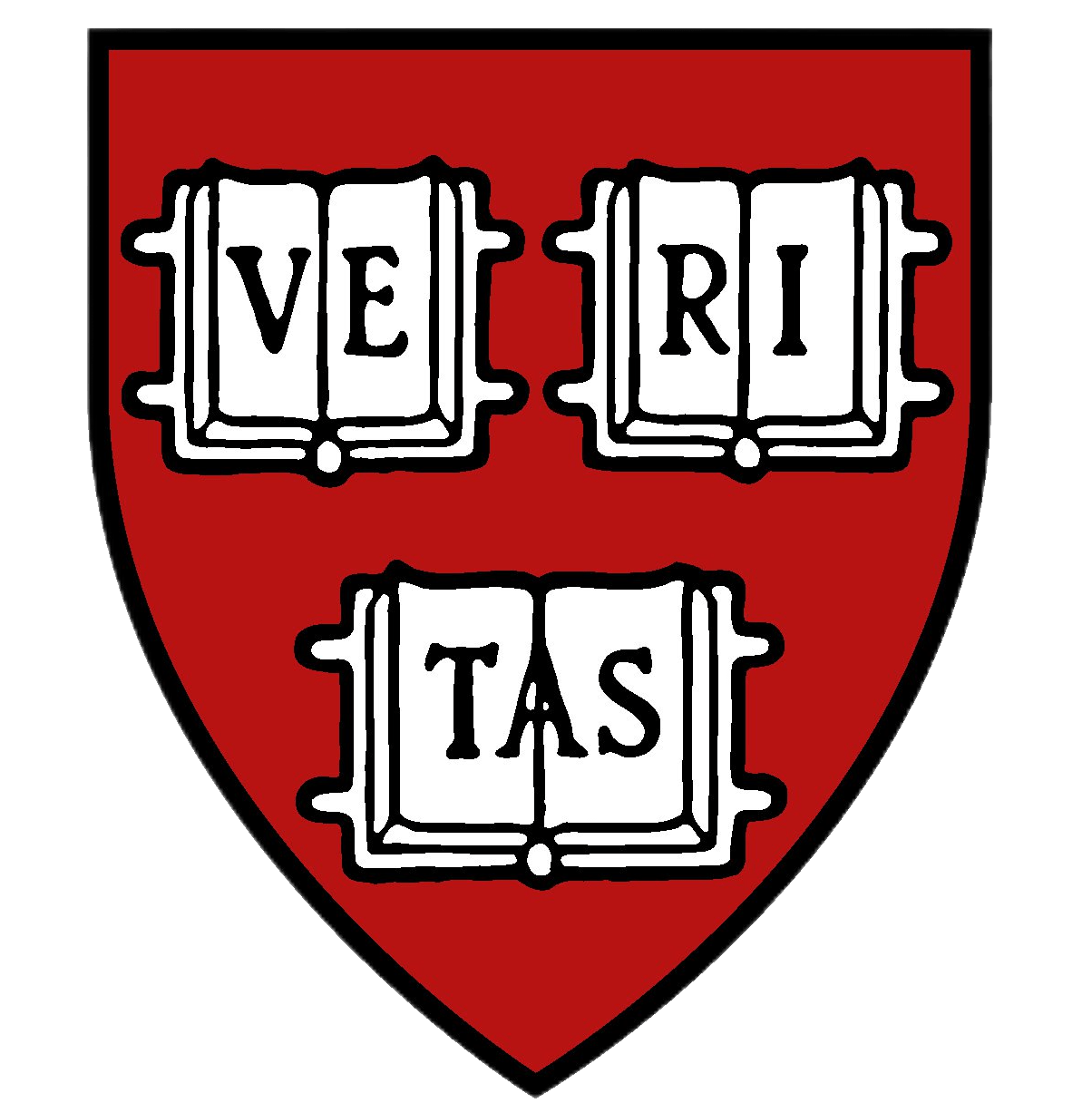}}}
\newcommand{\uclalogo}{\raisebox{3.4pt}{\includegraphics[scale=0.05]{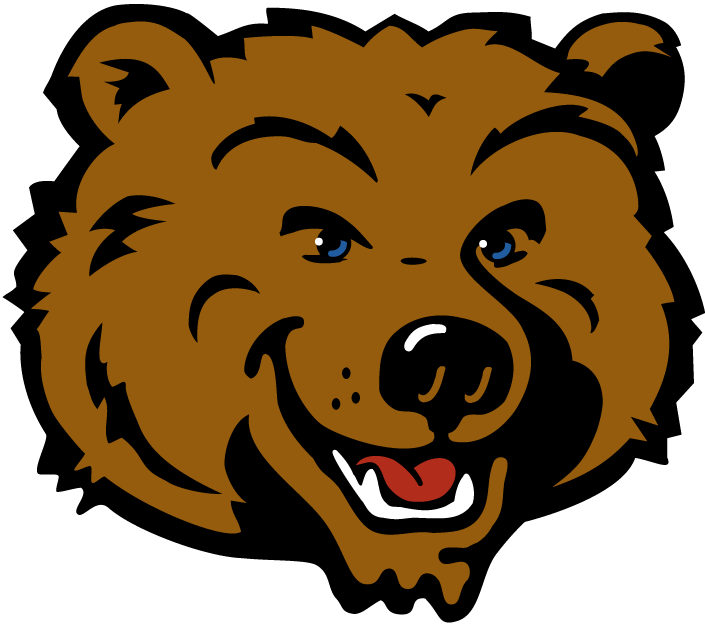}}}
\newcommand{\gtlogo}{\raisebox{3.4pt}{\includegraphics[scale=0.025]{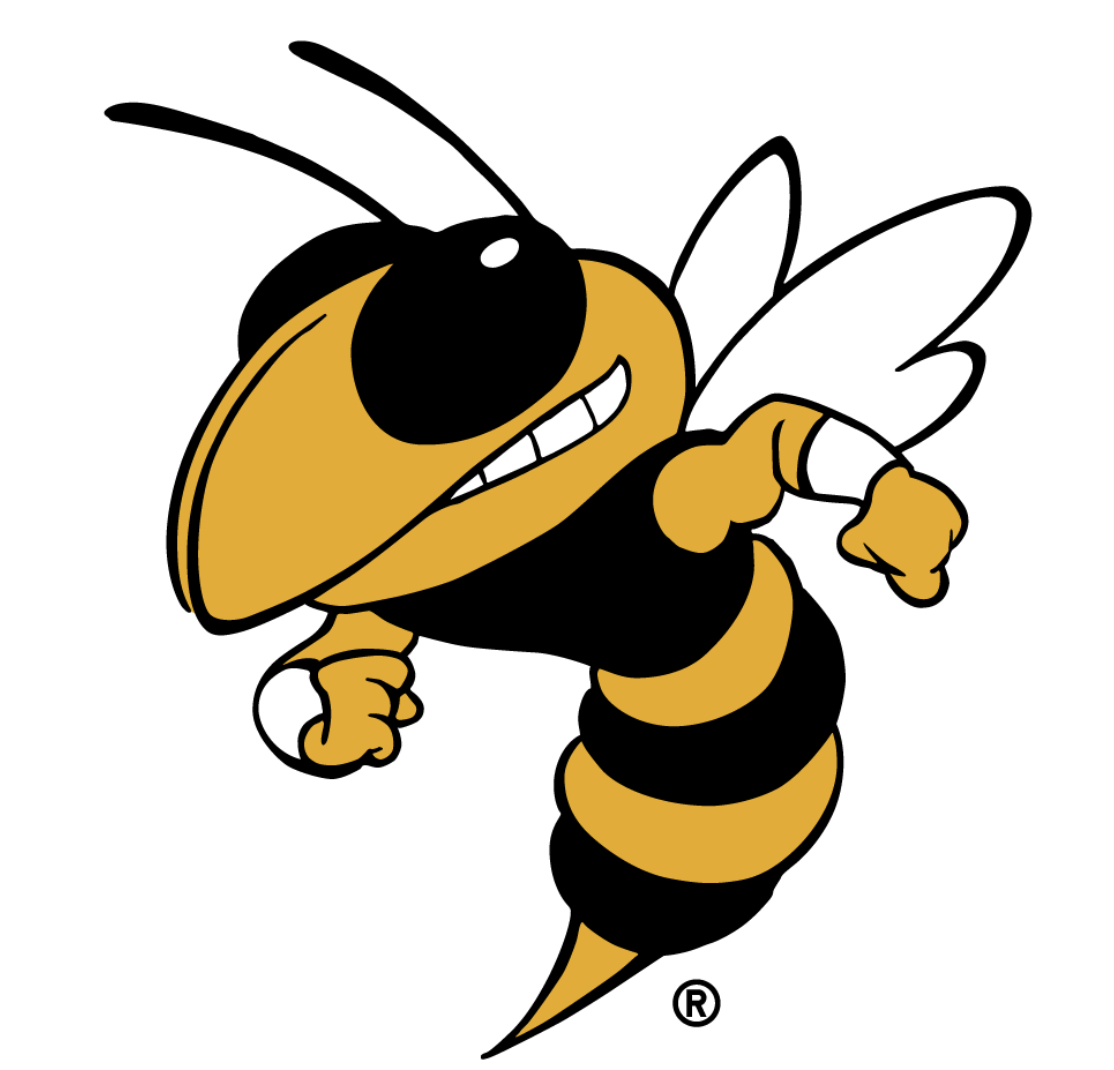}}}

\newcommand{\mypar}[1]{\paragraph{#1}}
\newcommand{\secref}[1]{\S\ref{#1}}

\newcommand{\figref}[1]{Fig.~\ref{#1}}
\newcommand{\tbref}[1]{Tab.~\ref{#1}}
\newcommand{\equationref}[1]{Eq.~\ref{#1}}

\newcommand{\vnew}[1]{\textcolor{black}{#1}}
\newcommand{\vn}[1]{\textcolor{black}{#1}}

\title{Decoding Susceptibility:
Modeling Misbelief to Misinformation Through a Computational Approach}

\author{Yanchen Liu\hlogo\hspace{2.5pt}
Mingyu Derek Ma\uclalogo\hspace{2.5pt}
Wenna Qin\treelogo\hspace{2.5pt}
Azure Zhou\treelogo\hspace{2.5pt}
Jiaao Chen\gtlogo\hspace{2.5pt}
\\
\textbf{Weiyan Shi\treelogo\hspace{2.5pt}
Wei Wang\uclalogo\hspace{2.5pt}
Diyi Yang\treelogo}
\\
\hlogo Harvard University \hspace{1pt}\treelogo Stanford University \hspace{1pt}\uclalogo UCLA \hspace{1pt}\gtlogo Georgia Institute of Technology\\
\texttt{yanchenliu@g.harvard.edu,} \texttt{\{wennaqin, amysz, weiyans, diyiy\}@cs.stanford.edu,} \\\texttt{\{ma,  weiwang\}@cs.ucla.edu}, \texttt{jiaaochen@gatech.edu}
}

\begin{document}
\maketitle
\begin{abstract}
    Susceptibility to misinformation describes the degree of belief in unverifiable claims, a latent aspect of individuals' mental processes that is not observable.
Existing susceptibility studies heavily rely on self-reported beliefs, which can be subject to bias, expensive to collect, and challenging to scale for downstream applications.
To address these limitations, in this work, we propose a computational approach to efficiently model users' latent susceptibility levels.
As shown in previous work, susceptibility is influenced by various factors (e.g., demographic factors,  political ideology), and directly influences people's reposting behavior on social media.
To represent the underlying mental process, our susceptibility modeling incorporates these factors as inputs, guided by the supervision of people's sharing behavior.
Using COVID-19 as a testbed, our experiments demonstrate a significant alignment between the susceptibility scores estimated by our computational modeling and human judgments, confirming the effectiveness of this latent modeling approach.
Furthermore, we apply our model to annotate susceptibility scores on a large-scale dataset and analyze the relationships between susceptibility with various factors. Our analysis reveals that political leanings and other psychological factors exhibit varying degrees of association with susceptibility to COVID-19 misinformation, \vn{and shows that susceptibility is unevenly distributed across different professional and geographical backgrounds}.\footnote{We will release all the code used in our paper, along with our trained model and all collected data. %
}

\end{abstract}

\begin{figure*}[!ht]
  \centering
  \includegraphics[width=\textwidth]{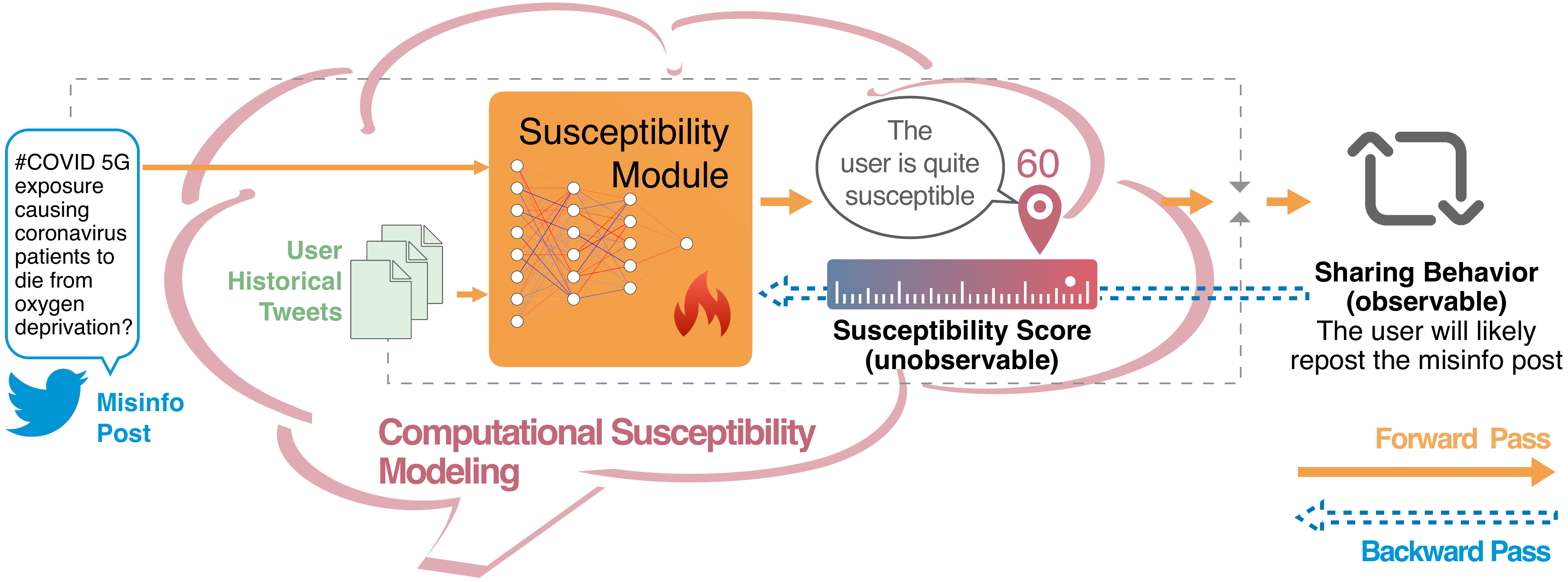}
  \caption{\textbf{Computational Modeling of Susceptibility to Misinformation}.
  We represent user susceptibility as a latent variable, which we capture using a shallow neural network.
  Our model is trained with the supervision of users'  observable sharing behaviors, employing two loss functions: \textit{binary classification entropy} and \textit{triplet loss}.
  }
  \label{fig:model}
  \vspace{-10pt}
\end{figure*}

\section{Introduction}
\label{sec:intro}

False claims spread on social media platforms, such as conspiracy theories, fake news, and unreliable health information. They mislead people's judgment, promote societal polarization, and exacerbate distrust in government~\cite{pennycook2021psychology, nan2021health}. The harm is especially significant in various contentious events, including elections, religious persecution, and the global response to the COVID-19 pandemic~\cite{ecker2022psychological}. 
Many works have investigated the \textit{observable} behavior of misinformation propagation such as where the information propagates~\cite{taylor2023does}, how people share it~\cite{yang2021covid}, and what people discuss about it~\cite{gupta2023polarised}. However, it is still crucial but challenging to understand the \textit{unobservable} mental and cognitive processes of how individuals believe misinformation~\cite{ecker2022psychological}. Individual susceptibility (i.e., the likelihood of believing and being influenced by misinformation) plays a pivotal role in this context. If one is more susceptible to misinformation, they are not only more likely to share but also more prone to being misled by them~\cite{scherer2021susceptible}.

Previous works have investigated the psychological, demographic, and other factors that may contribute to the high susceptibility of an individual~\cite{brashier2020aging, pennycook2020falls}.
However, these studies heavily rely on self-reported belief towards false claims collected from questionnaire-based participant surveys~\cite{escola2021critical, rosenzweig2021happiness}, which presents several limitations. For instance, different participants might interpret belief levels differently. Moreover, the data collection is labor-intensive, thereby limiting the scale of downstream research on the size, scope, and diversity of the target population~\cite{nan2022people}.

People's mental processes, which are unobservable and influenced by various factors, directly affect several externalized behaviors, such as reposting on social media~\cite{mitchell2019many, brady2020mad, islam2020misinformation, altay2022so}.
\vn{Building on these prior works, we propose a computational method to efficiently model individuals' unobservable susceptibility levels only based on their observable social media posting and sharing behaviors.
We represent users based on their historical posts and perform multi-task learning to simultaneously learn to classify whether a user would share a post, as well as to rank susceptibility scores among similar and dissimilar users when the same content is seen.}
This computational modeling method unlocks the scales of misinformation-related studies and provides a novel perspective to reveal users' belief patterns.

In this paper, we focus our experiments on COVID-19 misinformation, and our evaluations demonstrate that the estimations from our model are highly aligned with human judgment when assessed through a susceptibility comparison task. \vn{The correlation study between estimated and human-annotated susceptibility verifies the effectiveness of the indirect susceptibility modeling method.}
To further illustrate the significance of our work, we employ our model to annotate susceptibility levels on a large-scale dataset. Building upon this extensive susceptibility labeling, we then conduct a set analysis to examine how various factors relate to susceptibility.
Our analysis reveals that psychological factors, professional fields, and political leanings are associated with susceptibility to varying degrees. Notably, this large-scale analysis enabled by our computational susceptibility modeling corroborates the findings of previous studies based on self-reported beliefs, e.g. confirming that stronger analytical thinking is an indicator of lower susceptibility.
Moreover, the results of our analysis show the potential to extend findings in the existing literature. For example, we demonstrate that the distribution of COVID-19 misinformation susceptibility in the U.S. exhibits a certain degree of correlation with political leanings.

\section{Related Work}
\label{sec:related_work}

\mypar{Measure of Susceptibility} 
The common practice to measure susceptibility is to collect self-reported absolute or relative agreement or disagreement with (or perceived accuracy, credibility, reliability, or validity of) one or more claims verified to be false from a group of individuals~\cite{roozenbeek2020susceptibility, escola2021critical,rosenzweig2021happiness, nan2022people}.
A small number of previous studies indirectly assess the susceptibility by its impact, however, they can only capture behaviors rather than people's beliefs~\cite{loomba2021measuring}. \vnew{\citet{10.1145/3447548.3467321} defines a heuristic susceptibility score as the ratio of misinformation posts out of all user's posts, which unrealistically simplifies the definition of susceptibility.}
Instead of using expensive and limited self-reported beliefs, we propose a computational model to estimate susceptibility at scale.

\mypar{Contributing Factors and Application of Susceptibility}
Relying on the manually collected susceptibility annotation, previous research investigates the psychological, demographic, and more factors that contribute to users' susceptibility~\cite{bringula2022gullible,van2022misinformation}. These factors include emotion~\cite{sharma2023systematic} ~\citep[e.g. anger and anxiety;][]{weeks2015emotions}, analytic thinking~\cite{li2022emotion}, partisan bias~\cite{roozenbeek2022susceptibility}, source credibility~\cite{traberg2022birds}, and repetition~\cite{foster2012repetition}. Many theories have been proposed about the reason behind suscetibility~\cite{scherer2021susceptible}, including limited knowledge acquiring and literacies capabilities~\cite{brashier2020aging}, strong preexisting beliefs~\cite{lewandowsky2012misinformation}, neglecting to sufficiently reflect about the truth~\cite{pennycook2020falls} or overconfidence~\cite{salovich2021can}.\looseness=-1
A better understanding of the phenomenon and mechanism of susceptibility can facilitate various downstream applications. These include analyzing the spread of bots~\cite{himelein2021bots}, revealing community properties in information pathways~\cite{taylor2023does, ma2023middag}, combating misinformation by emphasizing publisher~\cite{dias2020emphasizing} and prebunking interventions based on inoculation~\cite{roozenbeekprebunking, roozenbeek2022psychological}.
However, the absence of a computational modeling framework significantly limits the scale of current susceptibility research.

\mypar{Inferring Unobservables from Observables}
Latent constructs or variables refer to concepts that are not directly observable or measurable.
Many studies have shown that unobservable variables can be inferred indirectly through models based on observable ones~\cite{bollen2002latent, borsboom2003theoretical}.
These unobservable variables can be estimated using various modeling techniques, including nonlinear mixed-effects models, hidden Markov models, or latent class models. 
In our work, we utilize a neural network-based architecture to model people's latent susceptibility level to misinformation, guided by the supervision provided by their observable sharing behaviors on social media.
\section{Computational Susceptibility Modeling}
\label{sec:method}

Misinformation is characterized as information that is false, inaccurate, or misleading, which could be created deliberately or accidentally~\cite{pennycook2021psychology}. The susceptibility to misinformation represents the belief in misinformation and related constructs, including discernment between true and false claims and the extent to which exposure to misinformation misleads subsequent decisions~\cite{nan2022people}.
\vn{Previous research on susceptibility and misinformation mainly relied on self-reported beliefs collected using surveys or questionnaires - they suffered from problems like being subject to bias, expensive to collect, and challenging to reproduce and scale up.}

\vn{Existing studies indicating that believing a piece of misinformation can influence various outward behaviors, such as sharing actions. For example, previous studies of the inattention or ``classical reasoning'' account contend that people are committed to sharing accurate information, but the unique context of social media disrupts their capacity to critically assess the accuracy of news \cite{pennycook2021psychology, van2022misinformation}. These studies suggest that people are more likely to share things they genuinely believe~\cite{altay2022so}.
Inspired by this observation, we propose to model user's unobservable susceptibility only based on their historical posting and sharing behaviors, which are the most available and the easiest collectable data from social media (\secref{sec:sus_model}) as shown in \figref{fig:model}. Therefore, our proposed framework can efficiently infer users' susceptibility levels to misinformation on a large scale, demonstrating the potential to expand the scope of previous misinformation-related research.}

\vn{Furthermore, because social media users utilize posts to express their personal and inner thoughts, they reveal information about their characteristics through their posts. Therefore, our proposed susceptibility modeling can incorporate users' informative hidden factors, such as personality traits, analytical thinking, and emotion, to infer a user's susceptibility to misinformation. These additional pieces of information are otherwise very difficult to directly collect on social media.}

\subsection{Modeling Unobservable Susceptibility}
\label{sec:sus_model}
\mypar{Content-Sensitive Susceptibility}
In our work, we consider the susceptibility of user $u$ when a particular piece of misinformation $p$ is perceived (i.e. $s_{u, p}$).
This allows us to account for the fact that an individual's susceptibility can vary across different content, influenced by factors such as topics and linguistic styles. By focusing on the susceptibility to specific pieces of misinformation, we aim to create a more nuanced, fine-grained, and accurate representation of how users interact with and react to different misinformation.

\mypar{User and Misinfo Post Embeddings}
\vn{We induce user and post embeddings to reflect hidden factors of the user personality traits and content of the post.}
As a component of the computational model, we use SBERT~\cite{reimers-gurevych-2019-sentence}, which is developed upon RoBERTa-large~\cite{liu2019roberta}, to compute the embedding vector to represent the information contained in the misinformation and user historical posts.
We consider the misinformation post as a sentence and produce its representation with SBERT.
For the user embedding, we calculate the average of sentence representations for the user's recent original posts.
More specifically, for every user-post pair $(u, p)$, we gather the historical posts written by user $u$ within a 10-day window preceding the creation time of the misinformation post $p$, to learn a representation of user $u$ at that specific time.\footnote{We chose the 10-day timeframe because it provides a substantial amount of data to represent a user and is also recent enough to capture their dynamics.}

\mypar{Computational Model for Susceptibility} Given the input of user historical posts for the user $u$ and the content for misinformation post $p$, the susceptibility computational model is expected to produce the \textit{susceptibility score} $s_{u, p}$ as shown in \equationref{eq:sus}, reflecting the susceptibility of $u$ when $p$ is perceived. 

\begin{equation}\label{eq:sus}
    s_{u, p} = suscep(E(u), E(p))
\end{equation}

We first obtain the embeddings $E(p)$ and $E(u)$ for post $p$ and user $u$, where $u$ is represented by the user's historical tweets and $E$ is the frozen SBERT sentence embedding function. The susceptibility score is calculated by the function $suscep$, which is implemented as a multi-layer neural network, taking the concatenation of the user and post embeddings as inputs. During the training phase, we maintain the sentence embedder as a fixed component and exclusively train the weights for the $suscep$ function. Then the learned $suscep$ function can be applied to generate susceptibility scores for new pairs of users ($u$) and posts ($p$) during the inference process.

\mypar{Scale and Interpretation of Susceptibility Score} Furthermore, for better interpretability, we normalize the resulting susceptibility scores within the range of -100 to 100 using \textit{Min-Max} normalization. We define -100 to indicate that the individual holds the most resistance to misinformation, while 100 means the individual is easiest to believe in misinformation when encountered.

\subsection{Training with Supervision from Observable Behavior}
\label{sec:training}

Susceptibility is a latent variable and cannot be directly observed. Consequently, it is impractical to directly apply supervision to $s_{u, p}$ since only the user $u$ themselves know their own beliefs regarding content $p$.
To address this challenge, we regard susceptibility as a crucial factor for sharing behavior and train the susceptibility computational model using the supervision signals obtained from the observable behavior of sharing misinformation.

To determine the probability of user $u$ sharing post $p$, we compute the dot product of the embeddings of the user and post content, incorporating the susceptibility score for the same pair of $u$ and $p$ estimated by our model as a weighting factor, and pass the resulting value through a sigmoid function, as illustrated in \eqref{eq:retweet}.

\begin{equation}\label{eq:retweet}
    p_{\text{rp}} = \sigma \left( E(u) \cdot E(p) \cdot s_{u, p} \right)
    \vspace{-7pt}
\end{equation}

It is important to highlight that we do not directly utilize the \textit{susceptibility score} to estimate sharing probability because sharing behavior depends not solely on susceptibility levels but also on various potential confounding factors. For instance, it is possible that a user may possess a significantly high susceptibility score for a piece of misinformation but decides not to share it, potentially influenced by factors such as their personality, the impact of social influence, concerns about potential repercussions, and their emotional state at that specific moment, among other variables.
To account for these potential confounding factors as comprehensively as possible, we incorporate a dot product of the user and post embeddings into our model.

\begin{figure*}[!ht]
\begin{align}
\mathcal{L}_{\text{bce}}(u_i, p) &= -\left( y_i \log(p_{\text{rt}}(u_i, p)) + (1 - y_i) \log(1 - p_{\text{rt}}(u_i, p)) \right) \nonumber \\ 
\mathcal{L}_{\text{triplet}}(u_a, u_s, u_{ds}, p) &= \text{ReLU}(\Vert s_{u_{a},p} - s_{u_{s},p}\Vert_2^2 - \Vert s_{u_{a},p} - s_{u_{ds},p} \Vert_2^2 + \alpha) \nonumber \\
\mathcal{L}(u_a, u_s, u_{ds}, p) &= \frac{\lambda}{3} \sum_{i \in \{a, s, ds\}} \mathcal{L}_{\text{bce}}(u_i, p) + (1 - \lambda) \mathcal{L}_{\text{triplet}}(u_a, u_s, u_{ds}, p) \label{eq:loss}
\end{align}
\vspace{-15pt}
\label{fig:loss}
\end{figure*}

\vspace{-8pt}
\mypar{Objectives}
To better train our computational model, we perform multi-task learning to utilize different supervision signals.
First, we consider a binary classification task of estimating repost or not with a cross-entropy loss.
Additionally, we perform the triplet ranking task~\cite{chen2009ranking,hoffer2015deep} to distinguish the subtle differences among the susceptibility scores of multiple users when the same false content is present.

During each forward pass, our model is provided with three user-post pairs: the anchor pair $(u_a, p)$, the similar pair $(u_s, p)$, and the dissimilar pair $(u_{ds}, p)$.
We regard the similar user $u_s$ as the user who reposted $p$ if and only if user $u_a$ reposted $p$. The dissimilar user $u_{ds}$ is defined by reversing this relationship. When multiple candidate users exist for either $u_s$ or $u_{ds}$, we randomly select one. However, if there are no suitable candidate users available, we randomly sample one from the positive (for ``reposted'' cases) or negative examples (for ``did not repost'' cases) and pair this randomly chosen user with the misinformation post $p$.

In \equationref{eq:loss}, we define our loss function. Here, $y_i$ takes the value of 1 if and only if user $u_i$ reposted misinformation post $p$. The parameter $\alpha$ corresponds to the margin employed in the triplet loss, serving as a hyperparameter that determines the minimum distance difference needed between the anchor and the similar or dissimilar sample 
for the loss to equal zero. Besides, $\lambda$ is the control hyperparameter, which governs the weighting of the binary cross-entropy and triplet loss components.
\section{Dataset and Experiment Setup}
\label{sec:data_training}
\vn{We have chosen Twitter as our data source because it hosts a diverse collection of users and allows for free-text personal and emotional expression. Furthermore, Twitter provides crucial metadata, including timestamps and location data, which are useful for our subsequent analysis.}

\mypar{Misinformation Tweets}
We consider two misinformation tweet datasets: the \texttt{ANTi-Vax} dataset \cite{hayawi2022anti} was collected and annotated specifically for COVID-19 vaccine misinformation tweets. And \texttt{CoAID} \citep[Covid-19 Healthcare Misinformation Dataset;][]{cui2020coaid} encompasses a broader range of misinformation related to COVID-19 healthcare, including fake news on websites and social platforms. The former dataset contains 3,775 instances of misinformation tweets, while the latter contains 10,443.
However, a substantial number of tweets within these two datasets do not have any repost history. Hence, we choose to retain only those misinformation tweets that have been retweeted by valid users.
Finally, we have collected a total of 1,271 misinformation tweets for our study.

\mypar{Positive Examples}
We define the positive examples for modeling as $(u_{pos}, p)$ pairs, where user $u_{pos}$ viewed and retweeted the misinformation post $p$. We obtained all retweeters for each misinformation tweet through the Twitter API.

\vspace{-0.5em}
\mypar{Negative Examples}
Regarding negative examples, we define them as $(u_{neg}, p)$ pairs where user $u_{neg}$ viewed but did not retweet misinfo post $p$. However, obtaining these negative examples poses a considerable challenge, because the Twitter API does not provide information on the ``being viewed'' activities of a specific tweet.
To address this issue, we construct potential negative users $u_{neg}$ who are highly likely to have viewed a particular post $p$ but did not repost it, following these heuristics: 1) $u_{neg}$ should be a follower of the author of the misinformation post $p$, 2) $u_{neg}$ should not retweet $p$, and 3) $u_{neg}$ was active on Twitter within 10 days before and 2 days after the timestamp of $p$.

In the end, we collected 3,811 positive examples and 3,847 negative examples, resulting in a dataset consisting of a total of 7,658 user-post pairs. 
We divide the dataset into three subsets with an 80\% - 10\% - 10\% split for train, validation, and test purposes, respectively.
The detailed statistics of the collected data are illustrated in \tbref{tab:data_statistics}.
We provide the training details of our model in Appendix \ref{appendix:training}.

\begin{table}[!t]
    \centering
    \resizebox{0.40\textwidth}{!}{%
    \begin{tabular}{l|c|cc}
        & Total & Positive & Negative \\\hline
    \# Example & 7658 & 3811 & 3847 \\\hdashline
    \# User & 6908 & 3669 & 3255 \\
    \# Misinfo tweet & 1271 & 787 & 1028 
    \end{tabular}}
    \caption{\textbf{Data Statistics} of our constructed training dataset. We show the statistics for the number of the user-tweet pairs (\textit{\# Example}), unique users (\textit{\# User}), and unique misinformation tweets (\textit{\# Misinfo tweet}) in the overall dataset and the positive and negative subsets.}
    \label{tab:data_statistics}
    \vspace{-15pt}
\end{table}
\section{Evaluation}
\label{sec:eval}

We demonstrate the effectiveness of our susceptibility modeling by directly comparing our estimations with human judgment (\secref{subsec:human_eval}) and indirectly evaluating for assessing sharing behavior (\secref{subsec:retweet_dist}, \secref{subsec:retweet_eval}).

\subsection{Validation with Human Judgement}
\label{subsec:human_eval}

Due to the abstract nature of susceptibility and the absence of concrete ground truth, we encounter challenges in directly assessing our susceptibility modeling. As a result, we tend to human evaluations to validate the effectiveness of our modeled susceptibility.
Given the inherent subjectivity in the concept of susceptibility, and to mitigate potential issues arising from variations in individual evaluation scales, we opt not to request humans to annotate a user's susceptibility directly.
Instead, we structure the human evaluation as presenting human evaluators with pairs of users along with their historical tweets and requesting them to determine which user appears more susceptible to overall COVID-19 misinformation.
We provide more details regarding the human judgment framework and the utilized interface in  Appendix \ref{appendix:interface}.

Subsequently, we compared the predictions made by our model with the human-annotated predictions. To obtain predictions from our model, we compute each user's susceptibility to overall COVID-19 misinformation by averaging their susceptibility scores to each COVID-19 misinformation tweet in our dataset.
As presented in Tab. \ref{tab:human_eval}, our model achieves an agreement of 72.90\% with human predictions, indicating a solid alignment with the annotations provided by human evaluators. Additionally, we consider a baseline that directly calculates susceptibility scores as the cosine similarity between the user and misinformation tweet embeddings. Compared to this baseline, our susceptibility modeling brings a 9.35\% improvement. 
Moreover, we conduct a comparison with ChatGPT by providing it with instructions based on the task description of the susceptibility level comparison setting in a zero-shot manner (more details are in Appendix \ref{appendix:prompt}). We notice that our model even outperforms predictions made by ChatGPT, despite ChatGPT being a significantly larger model than ours.
These results of the human judgment validate the effectiveness of our proposed susceptibility modeling, showcasing its capability to reliably estimate user susceptibility to COVID-19 misinformation.

\begin{table}[!ht]
    \centering
    \resizebox{0.4 \textwidth}{!}{%
    \begin{tabular}{c|ccc}
        & Our & Baseline & ChatGPT \\\hline
    Agreement & $72.90$ & $63.55$ & $62.62$
        
    \end{tabular}}
    \caption{\textbf{Comparison with Human Judgement}. \texttt{Baseline} refers to a direct comparison based on cosine similarity between user and misinformation embeddings, while \texttt{ChatGPT} denotes prompting the ChatGPT model (engine \textit{gpt-3.5-turbo-1106}) for determining the more susceptible user in a zero-shot manner.
    \vspace{-15pt}}
    \label{tab:human_eval}
\end{table}

\subsection{Inferred Susceptibility Score Distribution}
\label{subsec:retweet_dist}
\begin{figure}[!t]
  \centering
  \includegraphics[width=\linewidth]{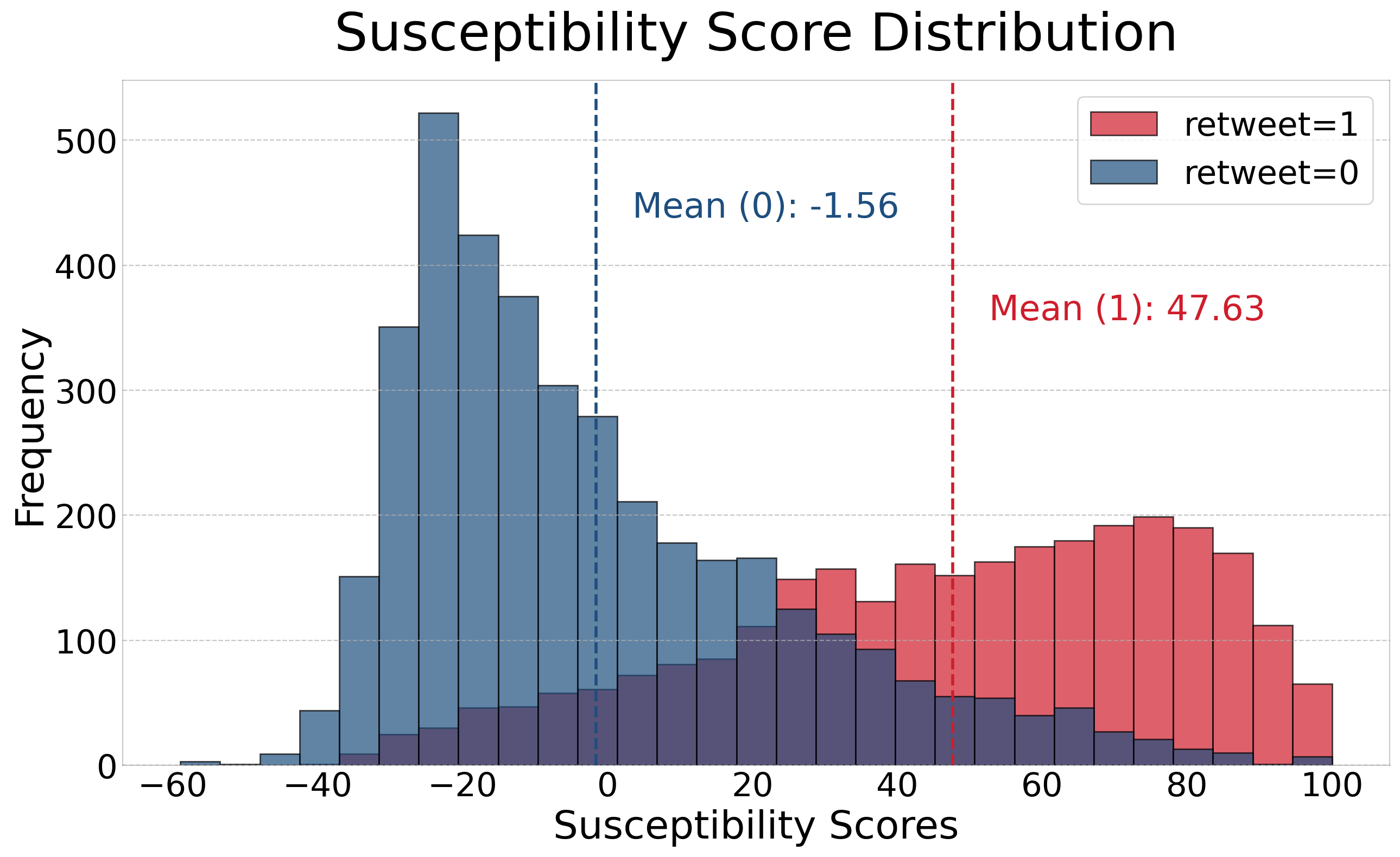}
  \caption{\textbf{Susceptibility Score Distribution} among positive and negative user-tweet pairs. The distribution of susceptibility levels, estimated by our computational modeling, among positive (\textcolor{myred}{red}) and negative (\textcolor{myblue}{blue}) examples exhibits a significant difference.}
  \label{fig:pos_neg_distribution}
  \vspace{-13pt}
\end{figure}

We provide a visualization showing the distribution of susceptibility scores produced by our model for both the positive and negative examples within the training data.
As illustrated in \figref{fig:pos_neg_distribution}, there is a significant disparity in the distribution between positive and negative examples. The difference in the means of the positive and negative groups is statistically significant, with a p-value of less than 0.001.
This confirms our assumption that the susceptibility level to misinformation is a fundamental influencing factor for subsequent sharing behavior.

\subsection{Resulting Sharing Behavior Prediction}
\label{subsec:retweet_eval}
Additionally, as described in \secref{sec:method}, a high susceptibility level to misinformation is highly likely to lead to subsequent sharing behavior on social media.
Here, we reinforce this assumption by showcasing that our learned susceptibility model exhibits a strong capability to predict subsequent sharing behavior.
When tested on the held-out test set, our model achieves a test accuracy of 78.11\% and an F1 score of 77.93.
These results indirectly demonstrate the validity of our computational modeling for latent susceptibility within the human thought process.


\section{Analysis}
\label{sec:analysis}

\vn{To further illustrate the significance of our work for the Computational Social Science community in susceptibility and misinformation research, we conducted a large-scale analysis on our collected large Twitter datasets and analyzed the correlation between user’s susceptibility and their psychological factors (\secref{subsec:corr}), professional backgrounds (\secref{subsec:community}), and geographical distribution (\secref{subsec:community}). Our findings demonstrate that the large-scale analysis enabled by our proposed efficient susceptibility modeling not only corroborates the results of previous questionnaire-based studies, but also shows the potential of further extending the scope of research on susceptibility and misinformation.}

\subsection{Correlation with Psychological Factors}
\label{subsec:corr}

Previous research on human susceptibility to health and COVID-19 misinformation primarily relied on questionnaire surveys \cite{scherer2021susceptible, nan2022people, van2022misinformation}. These studies have identified several psychological factors that influence individuals' susceptibility to misinformation. For instance, analytical thinking (as opposed to intuitive thinking), trust in science, and positive emotions have been linked to a greater resistance to health misinformation. Conversely, susceptibility to health misinformation is frequently associated with factors such as conspiracy thinking, religiosity, conservative ideology, and negative emotions.
In this part, we analyze the correlation coefficients between our modeled susceptibility scores and the aforementioned factors to determine whether our results align with previous research findings.

To achieve this, we compute factor scores for each user in our dataset based on their historical tweets using LIWC Analysis.\footnote{\url{liwc.app}. For each user, we compute the final factor score by calculating the average value across the user's historical tweets. However, for emotional factors like anxiety and anger, which may appear less frequently, we choose to use the maximum value instead to better capture these emotions.} We mainly consider the following factors: \textit{Analytic Thinking}, Emotions (\textit{Positive} emotions, \textit{Anxious}, \textit{Angry} and \textit{Sad}), \textit{Swear}, \textit{Political Leaning}, \textit{Ethnicity}, \textit{Technology}, \textit{Religiosity}, \textit{Illness} and \textit{Wellness}. These factors have been extensively studied in previous works and can be inferred from a user's historical tweets.
We calculate and plot the Pearson correlation coefficients between each factor and the susceptibility level estimated by our model in \tbref{tab:corr}.

In our analysis, the correlations are consistent with findings from previous social science studies that relied on surveys to assess participants' health susceptibility. For instance, \textit{Analytic Thinking} is a strong indicator of low susceptibility, with a correlation coefficient of -0.31.
Conversely, certain features such as \textit{Swear}, \textit{Political Leaning}, and \textit{Angry} exhibit a weak correlation with a high susceptibility level.
These results not only corroborate the conclusions drawn from previous questionnaire-based studies \cite{van2022misinformation, nan2022people} but also provide further validation for the effectiveness of our computational modeling for susceptibility.

\begin{table}[t]
\centering
\resizebox{0.49 \textwidth}{!}{
    \begin{tabular}{lc|lc}
    Factors &  Coeff. & Factors &  Coeff. \\\hline
    \textcolor{myblue}{\textbf{Analytic Thinking}}    & -0.31       & Emotion - Positive   & -0.08       \\
    \textcolor{myred}{\textbf{Political Leaning}}    & 0.13        & Emotion - Anxious    & 0.08        \\
    Ethnicity            & 0.09        & \textcolor{myred}{\textbf{Emotion - Angry}}      & 0.16        \\
    Religiosity          & 0.10        & \textcolor{myred}{\textbf{Emotion - Sad}}        & 0.14        \\
    Technology           & -0.09       & \textcolor{myred}{\textbf{Swear}}                & 0.18        \\
    Illness              & 0.09        & Wellness             & -0.02       \\
    \end{tabular}}
    \caption{\textbf{Correlation Coefficients} between our modeled susceptibility levels and various psychological factors. Our model reveals correlations that are consistent with findings from prior questionnaire-based health susceptibility studies. The factors with absolute scores greater than 0.1 are highlighted in \textcolor{myred}{red} (+) and \textcolor{myblue}{blue} (-).}
    \label{tab:corr}
    \vspace{-12pt}
\end{table}

\subsection{Community Differences}
\label{subsec:community}

We further leverage our computational model to investigate how susceptibility level differs and compares between different community groups on social networks. Specifically, two different types of communities are considered: professional and geographical communities.

To perform a reliable analysis among different communities, a large-scale user dataset is needed. To address this requirement, we sample 100,000 users across the world from the existing COVID-19 Tweet Dataset \cite{taylor2023does} which contains all COVID-19-related tweets for a certain time\footnote{Besides, we make sure each sampled user has posted more than 100 historical tweets between January 2020 and April 2021. 
For each user, we utilize the Twitter API to gather their user descriptions and location information, after which we extract and categorize their occupations from their self-reported descriptions with ChatGPT in a zero-shot manner.}.
To obtain an aggregated susceptibility score for a community, we calculate the mean of individual susceptibility scores for all users within that community.

\mypar{Occupation and Professional Community}
We first explore how susceptibility varies among users with different occupations. 
There is a social consensus regarding the susceptibility of the practitioners within a specific occupation community.
For example, susceptibility scores towards health misinformation are expected to be significantly lower among experts in health-related fields compared to the general population \cite{van2022misinformation, nan2022people}.
We consider the following professional communities and compare their average susceptibility scores: Education (\textit{Edu}), Society and Public (\textit{S\&P}), Health and Medicine (\textit{H\&M}), Finance and Business (\textit{F\&B}), Science and Technology (\textit{S\&T}), Arts and Media (\textit{A\&M}), as well as \textit{N/A}  for Twitter users who do not specify their occupation in their user descriptions.

The results are presented in Tab. \ref{tab:occupation}. It is worth noting that occupations within the \textit{A\&M} area demonstrate comparatively higher susceptibility, possibly because of their greater exposure to misinformation and stronger emotional reactions.
In contrast, professions closely associated with \textit{S\&T}, \textit{F\&B}, \textit{H\&M}, \textit{S\&P}, and \textit{Edu} tend to exhibit lower susceptibility to COVID-19 misinformation.
These findings reinforce the previous conclusions that expertise and knowledge in relevant fields serve as protective factors against misinformation, especially for populations in the field of \textit{H\&M} \cite{nan2022people}. Surprisingly, we notice that the \textit{S\&T} group is the most susceptible among the unsusceptible groups. For this, we have some speculations about the underlying reasons; for example, some people in the \textit{S\&T} community might have a higher level of skepticism towards traditional institutions and expertise, partly influenced by the culture that values disruptive innovation\footnote{We notice that Twitter users who don't declare their occupation in their user description (\textit{N/A}) exhibit a higher susceptibility to COVID misinformation. This may be because those who are willing to declare their profession are often public figures who care more about their reputation.}.

\begin{table}[!ht]
    \centering
    \resizebox{0.42 \textwidth}{!}{
    \begin{tabular}{l|rr}
    Occupation             & Suscep.  & \# Users   \\
    \hline
    N/A                   & 4.6201  & 35145 \\
    \textcolor{myred}{\textbf{Arts and Media}}         & -0.1504 & 12635 \\
    Science and Technology & -2.2076 & 7170  \\
    Finance and Business   & -5.4192 & 5844  \\
    \textcolor{myblue}{\textbf{Health and Medicine}}    & -5.4762 & 6272  \\
    Society and Public     & -6.7747 & 10973 \\
    Education              & -7.8070 & 5261  \\
    \end{tabular}}
    \caption{\textbf{Susceptibility Distribution by Professional Field}. We present the average susceptibility scores, estimated by our computational modeling, for 6 main professional fields. \textit{H\&M} (highlighted in \textcolor{myblue}{blue}) tends to have lower susceptibility to COVID-19 misinformation, consistent with existing studies.}
    \label{tab:occupation}
    \vspace{-8pt}
\end{table}

\begin{figure}[!ht]
  \vspace{-3pt}
  \centering
  \includegraphics[width=\linewidth]{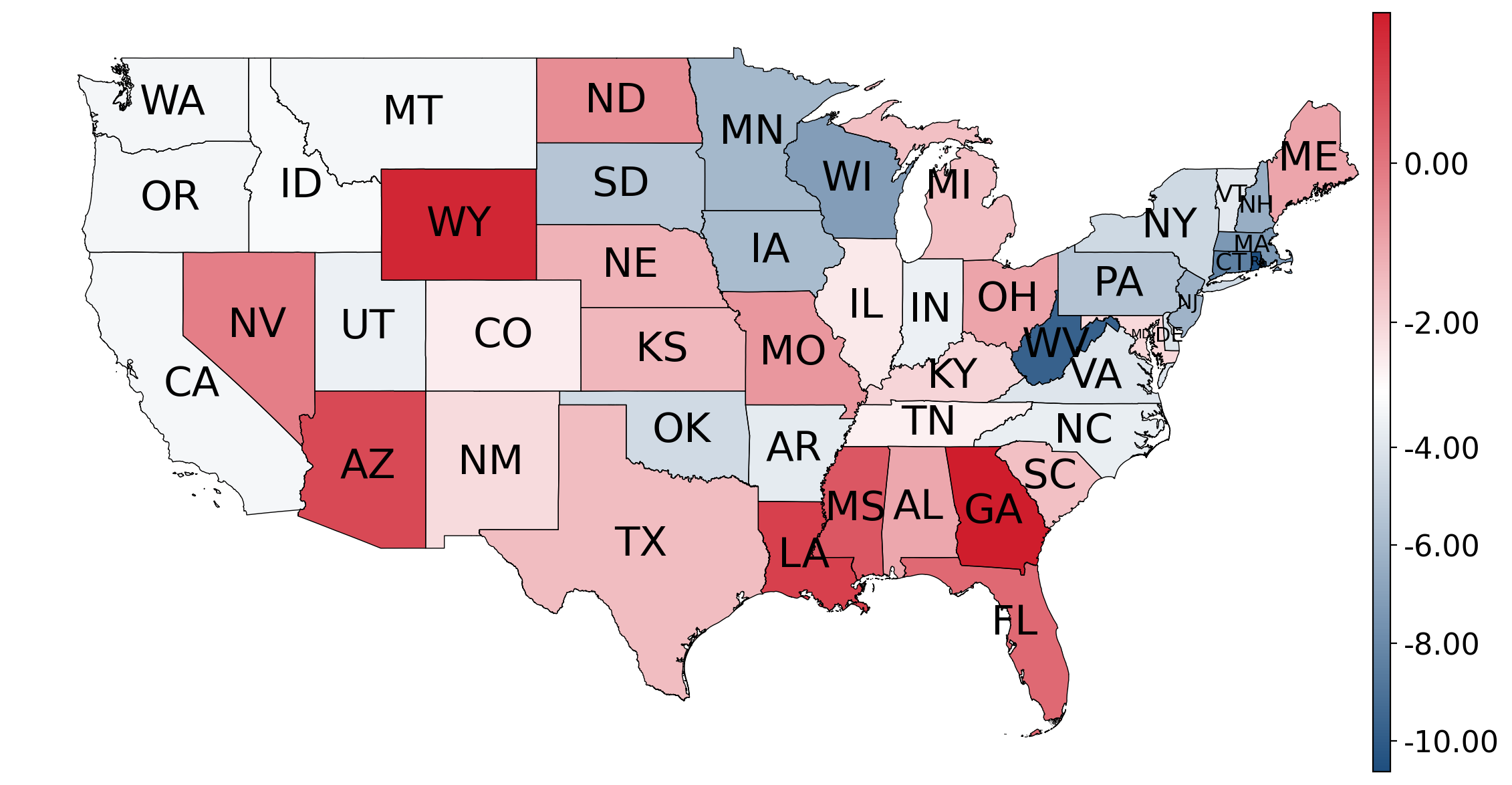}
  \caption{\textbf{Susceptibility Distribution by U.S. State}. We plot the susceptibility score, estimated by our computational modeling (with Bayesian smoothing), for each state in the U.S. The average susceptibility score in the overall U.S. (-2.87) is used as the threshold, with scores above it displayed in \textcolor{myred}{red}, and those below it in \textcolor{myblue}{blue}. Due to insufficient data points, we are only displaying data for 48 contiguous states within the U.S.}
  \label{fig:usa}
  \vspace{-8pt}
\end{figure}

\mypar{Geographical Community}
\label{par:geo_comm}
We further investigate the geographical distribution of susceptibility to COVID-19 misinformation, specifically focusing on different U.S. states.\footnote{Given the imbalance in the number of users from different U.S. states, we calculate average susceptibility scores for each state with Bayesian smoothing. We use the overall mean and overall standard deviation as priors, and the more users in the state, the less the overall mean will affect that state's score.}
This analysis enables us to explore the influence of political ideology associated with different U.S. states \cite{gelman2009red} on susceptibility to misinformation.
Out of the 100,000 users sampled from around the world, 25,653 users are from U.S. states with more than 200 users for each state. As shown in \figref{fig:usa}, the distribution of susceptibility levels estimated by our computational modeling is imbalanced across U.S. states and demonstrates a certain degree of correlation with political leanings. In general, states known to have a more conservative population tend to have relatively higher susceptibility scores, while states that are considered more liberal have lower scores. 
The average susceptibility score for users in blue or red states is -3.66 and -2.82 respectively.\footnote{Red and blue states are determined by the 2020 presidential election results, with red states leaning Republican and blue states leaning Democratic.}
We observe that 60\% of the ten states with the highest susceptibility scores are red states, and 90\% of the ten states with the lowest susceptibility scores are blue states.
This trend corresponds with the conclusion observed in various previous studies, where political ideology influences people's perspectives on scientific information \cite{mccright2013influence, baptista2021influence, imhoff2022conspiracy}.
However, it is important to acknowledge the limitations of our analysis, as it solely reflects the estimated susceptibility distribution of the sampled users within each state. In Appendix \ref{appendix:suscep_scores}, we present the average susceptibility scores calculated based on our sampled users for each U.S. state, along with the corresponding number of users.


\vspace{-1pt}
\section{Conclusion}
\vspace{-1pt}
\label{sec:conclusion}
In this work, we propose a computational approach to efficiently model people's latent susceptibility to misinformation.
While previous research on susceptibility is heavily relied on self-reported beliefs collected from questionnaire-based surveys, our model trained in a multi-task manner can estimate user's susceptibility levels only based on their posting and sharing behaviors on social media.
When compared with human judgment, our model shows highly aligned predictions on a susceptibility comparison evaluation task.
To demonstrate the potential of our proposed computational modeling in extending the scope of previous misinformation-related studies, we leverage the susceptibility scores estimated by our model to analyze factors that influence susceptibility to COVID-19 misinformation.
Our analysis considers a diverse population from various professional and geographical backgrounds, and the results obtained through our computational modeling not only align with but also support and extend the findings from previous survey-based social science studies.

\section*{Limitations}
\label{limitation}
Besides investigating the underlying mechanism of misinformation propagation at a large scale, the susceptibility scores estimated by our model have the potential to be used to visualize and interpret individual and community vulnerability in information propagation paths, identify users with high risks of believing in false claims and take preventative measures, and use as predictors for other human behaviors. However, while our research represents a significant step in computational modeling susceptibility to misinformation, several limitations should be acknowledged.

First, our model provides insights into susceptibility based on the available data and the features we have incorporated. However, it's important to recognize that various other factors, both individual and contextual, may influence susceptibility to misinformation. These factors, such as personal experiences and offline social interactions, have not been comprehensively incorporated into our modeling and should be considered in future research. 

Moreover, our modeled susceptibility scores represent an estimation of an individual's likelihood to engage with misinformation. These scores may not always align perfectly with real-world susceptibility levels. Actual susceptibility is a complex interplay of cognitive, psychological, and social factors that cannot be entirely captured through computational modeling.
Our modeling should be viewed as a valuable tool for identifying trends and patterns, rather than as a means for providing definitive individual susceptibility assessments.

Additionally, we employ correlational analysis to investigate the relationships between susceptibility to misinformation and various factors, professional and geographical backgrounds (\secref{sec:analysis}). It is crucial to note, however, that these correlations do not imply causation. For example, while our findings suggest an association between higher levels of analytical thinking and reduced susceptibility to misinformation, we cannot conclude that analytical thinking directly causes this low susceptibility. The results suggest potential relationships that are worth further investigation through causal research methods to explore the underlying mechanisms of these associations.

Finally, our study's findings are based on a specific dataset and may not be fully generalizable to all populations, platforms, or types of misinformation. Especially when examining the geographical distribution of susceptibility, it's important to note that not all U.S. states have a sufficient amount of Twitter data available for analysis, due to the high cost of data collection. Furthermore, platform-specific differences and variations in the types of misinformation can potentially impact the effectiveness of our modeling and the interpretation of susceptibility scores.

\section*{Ethics Statement}
Analyzing and modeling susceptibility to misinformation can potentially raise several ethical concerns, particularly when applied at an individual level. Due to its dual nature, our modeling can not only be used to identify users with a high risk of believing in misinformation and taking preventative measures to reduce harm, but it also holds the potential for misuse by malicious actors, leading to privacy violations, stigmatization, and targeted attacks.
To minimize the risk, we refrained from using any personally identifiable information (PII) data in our work. Nevertheless, it remains important to carefully consider the ethical implications associated with the deployment of computational models like ours, enhance regulatory oversight, and ensure responsible and transparent utilization. 

We acknowledge the need for ongoing ethical scrutiny and are committed to the responsible release of our trained model, and this includes requiring users to sign a Data Use Agreement that explicitly prohibits any malicious or harmful use of our model. Within this agreement, researchers and practitioners will also be required to acknowledge the limitations (\secref{limitation}), that our modeling may not fully or accurately represent an individual's real susceptibility level.

\section*{Acknowledgement}
We would like to thank the anonymous reviewers and the lab members from Stanford SALT and UCLA for their valuable feedback.
This work was sponsored by the Defense Advanced Research Project Agency (DARPA) grant HR00112290103/HR0011260656 and HR00112490370, the NSF grant IIS-2200274,  IIS-2106859 and IIS-2312501, as well as the NIH grant U54HG012517 and U24DK097771.

\bibliography{anthology,custom}

\newpage
\clearpage
\appendix
\section{\vn{Potential Questions}}
We address here some potential questions readers might have about our work:

\paragraph{What is the goal of the method design?}
We aim to design a framework to estimate users' susceptibility to misinformaion efficiently and scalably - indirectly modeling their susceptibility with a comprehensive representation of their observable reposting behavior data, rather than training on their ground-truth susceptibility levels. The sentence embedding model (described in Appendix~\ref{appendix:training}) is selected to create rich representations of users and posts. Its effectiveness has been shown in existing works \cite{levine2022the, liu-etal-2022-makes, liu-etal-2023-semantic, xu2023exploring, yang2024gpt4tools, deng2024mind2web}.

Moreover, the focus of our work is to develop a concise but reasonable framework for susceptibility modeling and demonstrate its effectiveness, rather than necessarily striving for the most optimal model. There is other information that could be used to model user susceptibility; however, most of it is not easy to collect and hence goes against our original motivation. We leave more complex signals for future work.

\paragraph{Proposed framework lacks novelty?}
\textit{Multi-fold novelty:} Our work contributes to the literature in multiple dimensions.
1) Proposing a brand new task without existing data, baselines and evaluation setup. Modeling user’s susceptibility efficiently and empirically while no ground-truth susceptibility to train or evaluate is provided;
2) Being the first to make large-scale susceptibility analysis possible, while previous works rely on expensive self-reported human-collected questionnaires;
3) Being the first large-scale analysis of the relationship between susceptibility and social/psych. factors, professional backgrounds and geographical distribution.

\textit{Conventional model is a secondary component under a bigger framework:} Even though we use RoBERTa model trained in previous works to obtain user and post embeddings, we are the first to design an indirect estimation framework for susceptibility from users' history. The off-the-shelf sentence embedding model is a secondary component and it can be replaced with other models, such as LLMs.

\textit{Not just method novelty:} The susceptibility modeling framework/method is only one part of our contribution, and more importantly, our other important core contribution is the large-scale analysis enabled by our proposed susceptibility modeling method and the interesting findings shown by this large-scale analysis. These findings not only corroborate the findings of previous questionnaire-based studies (which are not possible to scaled-up) but also showing the potential of extending the scope of misinfo research.

\paragraph{Reposting behavior not sufficient to provide a full understanding of believing?}
We acknowledge that modeling the user’s susceptibility to misinformation only with the supervision of their sharing behavior on social media is a little bit limited. However, other information, like ``user's intent behind reposting'', is almost never indicated on any social media, and intractable to large-scaly collect and impossible to scale up, which goes against our original motivation. And actually, only users themselves know their sharing intents, whether they are expressing approval or perhaps irony, which we believe are rare cases. Therefore, we propose to infer a user’s susceptibility to misinformation based solely on their historical tweets, because user’s historical posts and reposting behavior are much easier to collect. This not only enables effective modeling of user’s susceptibility and more importantly, it enables large-scale analysis to help people better understand the behind mechanism, patterns, influencing factors and distribution of human’s susceptibility to misinformation, which has been shown in our work. We have also acknowledged the data unavailability in the Limitations (\secref{limitation}) of this paper.


\paragraph{Why not use user personality features?}
We do not explicitly incorporate additional user characteristics into our modeling, because the additional information is relatively difficult to get on social media. However, users could display their personalities, etc., user characteristics information through their posts, thus justifying our design choice that our modeling solely based on users’ historical posts could also take these user characteristics into account.
To further confirm our point, there are lots of previous works proposing to predict/extract a user's personality from their posted or liked social media posts \cite{golbeck2011predicting, alsadhan2017estimating}.

\paragraph{How reliable are LIWC scores used in analysis in Section \ref{subsec:corr}?}
LIWC is a widely used, well-established, and convenient tool for analyzing text data in the field of computational linguistics and psychology. There are substantial works based on LIWC analysis and the reliability of LIWC has also been demonstrated in numerous studies across various domains \cite{wang2016twitter, chung2018we, sundararajan2022religion, boyd2022development}.

\paragraph{Why not add more comparing baselines?} 
We work on a brand new task setting: estimate users' susceptibility \textbf{indirectly} without any ground-truth susceptibility labels provided. The only input to the estimation model is users' historical posts. The unique setting prevents us from finding any prior works that follow this challenging setting, which makes it impossible for us to conduct a direct comparison with existing susceptibility works. Hence, we come up with two methods: cosine similarity between embeddings and ChatGPT. We also observe that cosine similarity is not a weak baseline, and it yields even better performance than ChatGPT.

\paragraph{Why not include more ablation studies?} 
Our work mainly focuses on developing a novel framework for susceptibility modeling and demonstrates its potential to enable large-scale analysis and facilitate susceptibility and misinformation-related research. Thus, we prioritize our emphasis on designing a reasonable modeling, rather than necessarily aiming for the optimal modeling. This is because, as previously stated, the unobservable nature and lack of ground truth for susceptibility prevent us from directly optimizing the modeling for susceptibility itself; instead, we can only do so for the indirect sharing predictions task. Consequently, ablation studies are of very limited significance in this context. We believe that including too many ablation studies could even deviate the audience's focus away from our research goals.

\paragraph{How to/what is the performance of adapting the proposed framework to other domains besides COVID-19?}
The advantage of our method is the capability to estimate susceptibility without the need for ground-truth user susceptibility labels. Using users' historical posts, target posts, and users' retweet behavior labels is sufficient to train the model. 
We will release our code, and people can try to robustness check it and extend it to more domains.

\section{Training Details}
\label{appendix:training}
We use the \textit{sentence-transformers/all-roberta-large-v1} model from \texttt{sentence-transformers} as our sentence embedder. Through grid search on learning rates ranging from 1e-5 to 5e-4 and $\lambda$ values from 0 to 1, we train our model using a learning rate of 3e-5, set the hyperparameter $\lambda$ to 0.9, and the margin $\alpha$ to 1 for 100 epochs on the training set, as detailed in \secref{sec:method}. Following the training process, we select the checkpoint with the lowest validation loss and proceed to evaluate its performance on the test set.

\section{Human Judgement}
\label{appendix:interface}
Here, we provide details about the human judgment framework utilized in our work.

\begin{figure*}[!ht]
  \centering
  \includegraphics[width=\textwidth]{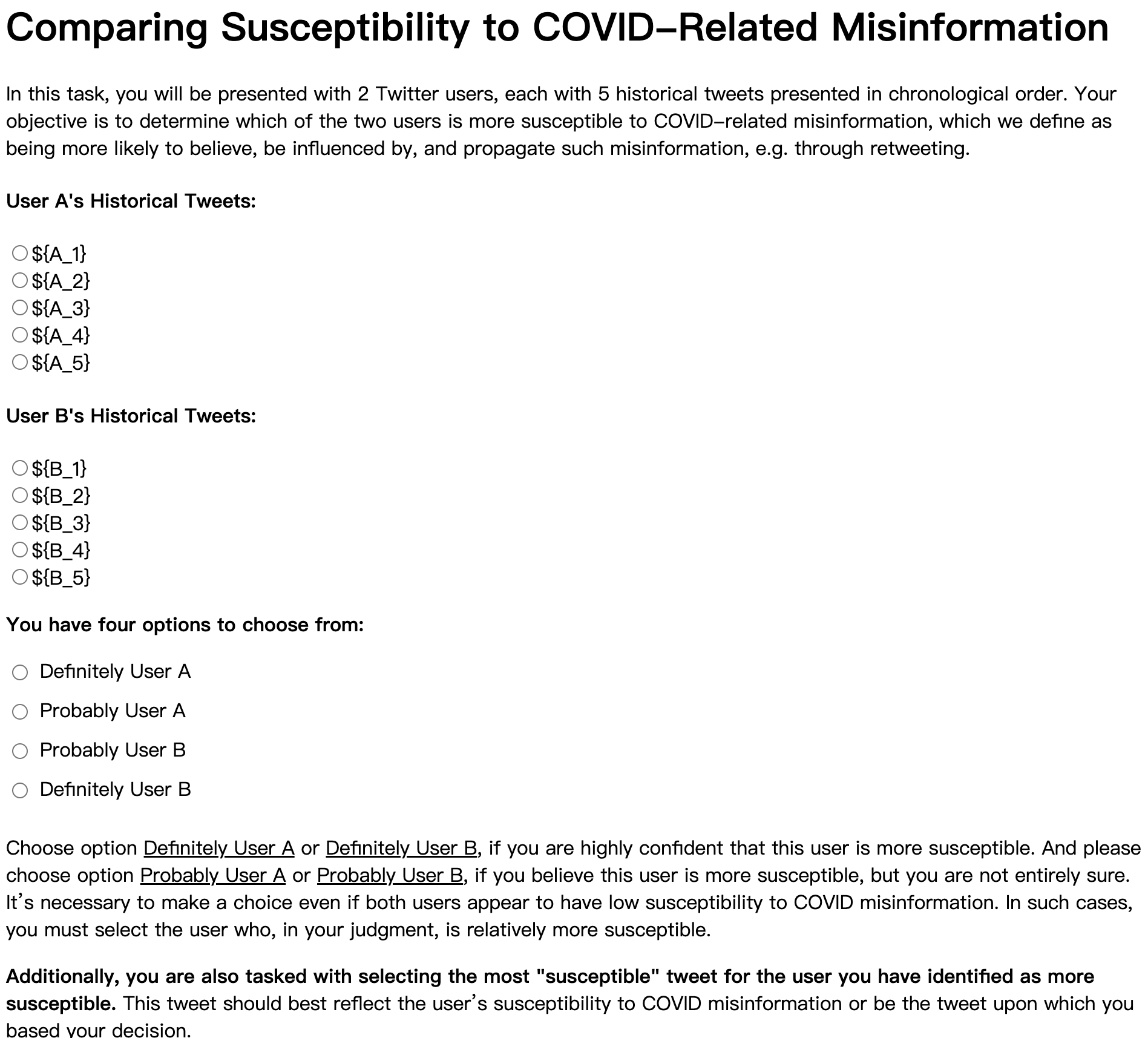}
  \caption{\textbf{Human Judgement Interface} utilized in our work. Participants are instructed to select the more susceptible user from a user pair based on five historical tweets for each user.}
  \label{fig:interface}
\end{figure*}

During human judgment, annotators are tasked with selecting the more susceptible user based on five historical tweets for each user. We offer the user interface used for human judgment in Figure \ref{fig:interface}.
In the task description, susceptibility is described as being more likely to believe, be influenced by, and propagate COVID-19 misinformation. 
To account for annotator uncertainty, we provide four options: \textit{Definitely User A}, \textit{Probably User A}, \textit{Definitely User B}, and \textit{Probably User B}. Furthermore, we also request annotators to identify the ``most susceptible tweet'' for the selected user, to enhance the reliability of annotations. This tweet should best exemplify the user's susceptibility to COVID-19 misinformation or be the basis for the annotator's decision.

Also, it is important to note that even when both users seem to have low susceptibility to COVID-19 misinformation, we still ask the annotator to make a choice. This is because our goal is to rank users based on their relative susceptibility, offering a comparative assessment rather than an absolute determination.

In total, we randomly sampled 110 user pairs and collected three annotations for each user pair. We recruited human annotators from Amazon Mechanical Turk (AMT) in the U.S. and compensated each annotator with \$0.5 (hourly wage higher than the federal minimum wage).
To determine the gold label for each user pair, we applied a weighted majority voting approach, assigning a value of 0.5 to \textit{Probably User X} and a value of 1 to \textit{Definitely User X}. We excluded user pairs with tied annotations, resulting in a final dataset of 107 user pairs. The kappa score for interrater agreement among the annotators is 0.74.

\section{Examples of User Posts and Susceptibility Scores}

Here we show some examples of users (characterized by their historical posts) and the susceptibility scores estimated by our model for each user when viewing a specific tweet. Please note that the target tweets we are showing here are randomly sampled COVID-19-related tweets, but they are not necessarily misinformation posts. This also suggests that our trained susceptibility model can be utilized to estimate a user's susceptibility  to both misinformation and non-misinformation.

The user \texttt{KatCapps}'s susceptibility score is estimated as \texttt{38.62} when the user sees the tweet:

\begin{tcolorbox} 
    \centering
    \begin{tabular}{p{0.94\columnwidth} c}
    The coronavirus infection rate is still too high. There will be a second wave | David Hunter \href{https://www.theguardian.com/commentisfree/2020/may/28/coronavirus-infection-rate-too-high-second-wave}{[Link]}
    \end{tabular}
\end{tcolorbox}

History tweets posted by the user are:

\begin{itemize}
    \item RT @gregggonsalves: Study estimates 24 states still have uncontrolled coronavirus spread
    \item RT @JoeSudbay: OSHA chas not issued enforceable guidelines for protecting employees from covid-19, as it did during the H1N1 outbreak in
    \item RT @mlipsitch: New Oped with @rickmalley | Treating Mild Coronavirus Cases Could Help Save Everyone - The New York Times
    \item RT @stevesilberman: Texas church that rushed to reopen cancels masses after priest dies and others contract \#coronavirus.
    \item RT @carlzimmer: Several cases of coronavirus reported after a swim party in Arkansas, governor says
    \item RT @GlennKesslerWP: They Survived the Worst Battles of World War II. And Died of the Virus.
\end{itemize}

Another user \texttt{AmitSin91018424}'s susceptibility score is estimated as \texttt{-12.27} when the user sees the tweet:

\begin{tcolorbox} 
    \centering
    \begin{tabular}{p{0.94\columnwidth} c}
    Dominic Cummings has broken Covid-19 policy trust, say top scientists \href{https://www.theguardian.com/politics/2020/may/30/dominic-cummings-has-broken-covid-19-policy-trust-say-top-scientists}{[Link]}
    \end{tabular}
\end{tcolorbox}

History tweets posted by the user are:

\begin{itemize}
    \item RT @guardian: The pandemic has laid bare the failings of Britain's centralised state | John Harris
    \item RT @guardian: Coronavirus world map: which countries have the most cases and deaths?
\end{itemize}

\section{ChatGPT Prompt Template}
\label{appendix:prompt}

In \figref{fig:chatgpt_prompt}, we present the template used to prompt ChatGPT for the susceptibility comparison task (\secref{subsec:human_eval}).

\begin{figure*}[!ht]{
\begin{lstlisting}[
    breaklines=true,
    postbreak=\mbox{\hspace{-0.7cm}},
    basicstyle=\color{myred}\ttfamily\footnotesize,
    columns=fullflexible,
    escapeinside={(*}{*)}
]
In this task, you will be presented with 2 Twitter users, each with 5 historical tweets presented in chronological order. Your task is to determine which of the two users is more susceptible to COVID-related misinformation, which we define as being more likely to believe, be influenced by, and propagate such misinformation, e.g. through retweeting.
\end{lstlisting}

\begin{lstlisting}[
    breaklines=true,
    postbreak=\mbox{\hspace{-0.7cm}},
    basicstyle=\color{myblue}\ttfamily\footnotesize,
    columns=fullflexible,
    escapeinside={(*}{*)}
]
User A's Historical Tweets:
{userA_text}

User B's Historical Tweets:
{userB_text}
\end{lstlisting}
}

\begin{lstlisting}[
    breaklines=true,
    postbreak=\mbox{\hspace{-0.7cm}},
    basicstyle=\color{myred}\ttfamily\footnotesize,
    columns=fullflexible,
    escapeinside={(*}{*)}
]
It is necessary to make a choice even if both users appear to have low susceptibility to COVID misinformation.
In such cases, you must select the user who, in your judgment, is relatively more susceptible.

Please answer with one of the following options without any other text: A | B.
\end{lstlisting}
\caption{\textbf{ChatGPT Prompt Template} for the susceptibility comparison task.}
\label{fig:chatgpt_prompt}
\vspace{-0.5em}
\end{figure*}

\vspace{-0.2em}
\section{Average Susceptibility Scores and User Counts by U.S. State}
\vspace{-0.2em}
\label{appendix:suscep_scores}

We provide the aggregated susceptibility scores estimated by our computational modeling for each U.S. state (\secref{par:geo_comm}), along with the number of sampled users in \tbref{tab:suscep_scores}.

\begin{table*}[!ht]
\centering
\begin{tabular}{l|rr||l|rr}
State & Suscep. & \# Users & State & Suscep. & \# Users \\
\hline
Georgia & 0.3935 & 669 & Idaho & -3.2296 & 265 \\
Florida & -0.2404 & 1592 & Washington & -3.2577 & 526 \\
Arizona & -0.5566 & 499 & Montana & -3.2590 & 543 \\
Louisiana & -1.3878 & 202 & Oregon & -3.2612 & 260 \\
Ohio & -1.6120 & 679 & Utah & -3.3324 & 206 \\
Texas & -1.7478 & 1627 & Vermont & -3.3548 & 556 \\
Missouri & -1.9076 & 308 & Indiana & -3.3901 & 270 \\
Nevada & -1.9857 & 294 & Delaware & -3.4139 & 359 \\
Michigan & -2.0996 & 575 & Arkansas & -3.4179 & 418 \\
Alabama & -2.3902 & 377 & North Carolina & -3.5324 & 635 \\
Maryland & -2.4763 & 527 & South Dakota & -3.6020 & 351 \\
South Carolina & -2.5456 & 298 & Virginia & -3.7276 & 528 \\
Mississippi & -2.5886 & 257 & Oklahoma & -3.7577 & 291 \\
Maine & -2.6193 & 208 & New Hampshire & -4.1011 & 399 \\
Illinois & -2.6294 & 816 & Iowa & -4.1603 & 249 \\
Nebraska & -2.6339 & 324 & New York & -4.4226 & 2835 \\
Kansas & -2.6541 & 328 & West Virginia & -4.8056 & 285 \\
Kentucky & -2.7774 & 469 & Minnesota & -4.8423 & 372 \\
Colorado & -2.8109 & 363 & Pennsylvania & -4.8700 & 873 \\
Tennessee & -2.8554 & 397 & Rhode Island & -5.0661 & 488 \\
New Mexico & -2.9178 & 518 & Wisconsin & -5.2446 & 279 \\
Wyoming & -2.9401 & 319 & New Jersey & -5.2594 & 598 \\
North Dakota & -2.9789 & 331 & Connecticut & -5.6912 & 242 \\
California & -3.2206 & 2849 & Massachusetts & -6.3191 & 761 \\
\end{tabular}
\caption{\textbf{Susceptibility Scores Estimated by Our Computational Model and Number of Sampled Users per U.S. State}. Due to insufficient data points, we only consider 48 contiguous states within the U.S.}
\label{tab:suscep_scores}
\end{table*}

\end{document}